\documentclass[letterpaper, 10pt, journal, twoside]{IEEEtran}
\newif\ifanonymous
\anonymousfalse
\usepackage[T1]{fontenc}
\usepackage{amsmath,amssymb,bm}
\usepackage{graphicx}
\usepackage{booktabs}
\usepackage{siunitx}
\usepackage{xcolor}
\usepackage{tikz}
\usetikzlibrary{arrows.meta,positioning,shapes.geometric,fit,calc,backgrounds}
\usepackage{cite}
\usepackage{microtype}
\usepackage{multirow}
\usepackage[hidelinks]{hyperref}

\title{RTISSLAM: In-air Imaging Sonar SLAM using a Robust Pose Graph and Occupancy Grid Mapping}
\ifanonymous
\author{Anonymous Authors}
\else
\author{Jan~Steckel%
\thanks{J. Steckel is with Cosys-Lab, Faculty of Applied Engineering, University of Antwerp, 2020 Antwerp, Belgium, and also with Flanders Make Strategic Research Centre, 3920 Lommel, Belgium (e-mail: jan.steckel@uantwerpen.be).}}
\fi
\ifanonymous\def\headauthors{Anonymous Authors}\else\def\headauthors{Steckel}\fi
\begin{document}
\maketitle

\begin{abstract}
Sonar was one of the first sensors with which mobile robots built maps, and in-air sonar keeps working where cameras and lidar fail, in darkness, smoke, dust and next to glass. Classical sonar SLAM, however, relied on geometric features or occupancy grids built from sparse range readings, and both struggle with the specular reflections that dominate indoor sonar. In this letter, we introduce RTISSLAM, a novel sonar-only SLAM system for an embedded 3D imaging sonar, which combines the robust pose graph of BatSLAM 2.0 as back-end wdfd~~~ith an occupancy grid as map. Energyscapes from delay-multiply-and-sum beamforming with a range-wise CFAR serve as place descriptors for sequence-verified loop closures, acoustic odometry calibrates the wheel odometry online, recognized revisits are registered metrically, and the grid is stored per keyframe and anchored to the pose-graph nodes, so that it follows every loop closure. In simulated worlds with specular walls, RTISSLAM reduces the trajectory error of the wheel odometry from \SI{8.2}{\meter} to \SI{0.31}{\meter} in an office, while lidar-style grid SLAM on the same sonar data fails. 
\end{abstract}


\section{Introduction}
\label{sec:intro}
Sonar is one of the oldest sensors that was widely used in mobile robotics. The very first occupancy grids were built from the range readings of a ring of Polaroid sonars \cite{moravecHighResolutionMaps1985, elfesUsingOccupancyGrids1989}, and the first feature-based localization and mapping systems tracked walls, corners and edges extracted from sonar data in an extended Kalman filter \cite{leonardMobileRobotLocalization1991, kleemanMobileRobotSonar1995, tardosRobustMappingLocalization2002}. Underwater, sonar has remained the sensor of choice for SLAM \cite{ribasUnderwaterSLAMManmade2008}. In air, however, it was replaced by laser scanners and cameras as soon as these became widely available and affordable. Still, in-air sonar keeps working in exactly those conditions in which optical sensors fail (darkness, smoke, dust and fog, and next to glass), and ultrasonic sensors are cheap, small and robust \cite{steckelBroadband3DSonar2013, kerstensERTISFullyEmbedded2019, Laurijssen2026eRTIS}.

While sonar sensing has its benefits in optically degraded conditions \cite{Laurijssen2026eRTIS}, the old and classical SLAM implementations for airborne sonar sensing has two main weaknesses. Firstly, it relied on transducers that measure only the range to the nearest reflector in a wide beam, so that a map had to be accumulated from many sparse and ambiguous readings, as explicit direction information was not available. Secondly, indoor surfaces reflect sound specularly: a wall is only seen near normal incidence, whereas corners, edges and door frames return echoes from almost every direction. Geometric features are therefore hard to extract, and occupancy grids built from such readings contain holes where walls were seen at grazing incidence, and phantom obstacles where multiple reflections were taken for direct ones \cite{moravecHighResolutionMaps1985, thrunProbabilisticRobotics2005}.

Recent developments in in-air sonar sensors make it worthwhile to revisit sonar SLAM. Firstly, embedded 3D imaging sonars such as the eRTIS \cite{kerstensERTISFullyEmbedded2019, Laurijssen2026eRTIS} beamform the signals of a N-channel microphone array into an energyscape, i.e., the reflected energy as a function of range and direction, for every pulse \cite{steckelBroadband3DSonar2013}. Nonlinear beamformers such as Delay-Multiply-And-Sum (DMAS) with a coherence factor have made these images considerably sharper, and can be efficiently implemented so they run in real time on embedded robotics GPUs \cite{Jansen2026DMAS}. Secondly, SLAM back-ends have matured into robust factor-graph optimizers \cite{kaessISAM2IncrementalSmoothing2012, dellaertFactorGraphsGTSAM2012} that withstand wrong loop closures \cite{sunderhaufSwitchableConstraintsRobust2012, agarwalRobustMapOptimization2013, yangGraduatedNonConvexityRobust2020}. The recently proposed BatSLAM 2.0 \cite{steckelBatSLAM20Sequenceverified2026} combined such a factor-graph back-end with sequence-verified place recognition on the echoes of a binaural bat-like sonar sensor, and produced reliable topological maps even in strongly aliased environments. While topological maps provide high-level path planning systems with usefull information, access to a true metric map at some scales would be beneficial. Recently, in-air sonar has also been used to estimate odometry and maps with generative models \cite{Jia2024GenerativeUltrasonicSLAM}, to map acoustic reflectors \cite{Qiu2025ParametricMapping, Saqib2024MLReflectorMapping, Saqib2024RobustEM}, alongside a lidar SLAM system \cite{Schenck2025AirLeaks}, and for the navigation of palm-sized aerial robots \cite{Velmurugan2026Saranga}? Furthermore, learned models predict sonar images \cite{Steckel2024EchoPT}, reconstruct scenes \cite{Jansen2024SonoNERF} or estimate depth from echoes \cite{Honma2024UltrasonicDepth, Liu2025SAGENet, Zhang2025EchoDiffusion}. This wide and renewed domain of applications sparked the interest in revisiting the occupancy grid SLAM systems of the past, to see if the new techniques can improve the construction of such occupancy grid maps using modern techniques. 

The question we address in this paper is how the spatial information of an imaging sonar is best combined with a robust pose graph. Our answer is a deliberate division of labor: the pose graph of BatSLAM 2.0 remains the back-end that estimates the trajectory of the robot, driven by place recognition on energyscapes. An occupancy grid map, anchored to the nodes of the pose graph, is the front-end that turns the energyscapes into a metric map. The pose estimate is thereby protected from the specular holes and phantom obstacles of sonar grids (due to the robust implementation of the pose graph using local view recognition), and the grid inherits the consistency of the pose graph after every loop closure. We deliberately use a pipeline without learned components, from the beamformer to the map, to make this system as widely applicable as possible. 

The contributions of this paper are as follows:
\begin{itemize}
\item RTISSLAM, a sonar-only SLAM system for an embedded imaging sonar, which uses DMAS energyscapes with a range-wise CFAR as place descriptors in the pose graph of BatSLAM 2.0, and an occupancy grid that is stored per keyframe and rendered from the current pose graph.
\item The use of acoustic odometry \cite{steckelBroadband3DSonar2013} with an exact flow warp to calibrate the wheel odometry online, which we show to be more effective than adding acoustic flow as an additional odometry factor.
\item The metric registration of recognized revisits on the echoes of both visits.
\item An evaluation in three simulated worlds with specular walls over five noise realizations, with an ablation and a comparison with lidar-style grid SLAM on the same sonar data.
\end{itemize}

\section{Overview of RTISSLAM}
\label{sec:overview}
\begin{figure*}[t]
\centering
\begin{tikzpicture}[
  box/.style={draw=black!70, rounded corners=2pt, align=center, font=\footnotesize, text width=2.45cm, minimum height=#1},
  box/.default=1.55cm,
  arr/.style={-{Stealth[length=2mm]}, thick, black!70},
  lab/.style={font=\scriptsize, black!65, align=center}]
\node[box, fill=blue!6] (echo) at (0,0) {eRTIS Pulse\\ $s_{M_i}[k]$, $i = 1 \ldots N$};
\node[box, fill=blue!6] (img) at (3.3,0) {Imaging\\ DMAS3 + CF\\ RW-CFAR, $P(r,\psi)$\\ (Section \ref{sec:imaging})};
\node[box, fill=blue!6] (lv) at (6.6,1.6) {Local View $V$\\ Templates\\ (Section \ref{sec:views})};
\node[box, fill=orange!10] (sv) at (9.9,1.6) {Sequence Verifier\\ (BatSLAM 2.0)};
\node[box, fill=orange!10] (reg) at (13.2,1.6) {Metric Registration\\ of Revisits\\ (Section \ref{sec:registration})};
\node[box, fill=blue!6] (ao) at (6.6,-0.9) {Acoustic Odometry\\ $(\Delta s_A, \Delta\theta_A)$\\ (Section \ref{sec:ao})};
\node[box, fill=blue!6] (cal) at (9.9,-0.9) {Wheel Calibration\\ $\kappa_s$, $\kappa_\theta$\\ (Section \ref{sec:calibration})};
\node[box, fill=green!8] (pg) at (13.2,-0.9) {Pose Graph\\ (iSAM2, link\\ management)\\ (Section \ref{sec:posegraph})};
\node[box=0.75cm, fill=white] (odo) at (9.9,-2.3) {Wheel Odometry $u_i$};
\node[box, fill=purple!8] (kf) at (6.6,-3.3) {Keyframe Evidence\\ $\mathcal{O}_k$, $\mathcal{F}_k$\\ (Section \ref{sec:grid})};
\node[box, fill=purple!8] (grid) at (13.2,-3.3) {Occupancy Grid\\ $l(\vec{m})$ rendered\\ from $\hat{X}$};
\draw[arr] (echo) -- (img);
\draw[arr] (img.east) -- ++(0.35,0) |- (lv.west);
\draw[arr] (img.east) -- ++(0.35,0) |- (ao.west);
\draw[arr] (img.east) -- ++(0.35,0) |- (kf.west);
\draw[arr] (lv) -- (sv);
\draw[arr] (sv) -- (reg);
\draw[arr] (reg) -- node[lab, right]{links $z_{qa}$} (pg);
\draw[arr] (ao) -- (cal);
\draw[arr] (odo) -- (cal);
\draw[arr] (cal) -- node[lab, below]{$K \cdot u_i$} (pg);
\draw[arr] (kf) -- node[lab, below]{anchored to node $a_k$} (grid);
\draw[arr] (pg) -- node[lab, right]{$\hat{X}$} (grid);
\draw[arr, dashed] ($(pg.north west)+(0.25,0)$) -- node[lab, above, sloped, pos=0.45]{estimate} ($(sv.south east)+(-0.25,0)$);
\end{tikzpicture}
\caption{Overview of the processing flow of RTISSLAM. Every pulse of the eRTIS is beamformed into an energyscape $P(r,\psi)$, which feeds the place recognition of BatSLAM 2.0 (local views, templates and the sequence verifier), the acoustic odometry and the keyframe evidence of the occupancy grid. The acoustic odometry calibrates the wheel odometry ($\kappa_s$, $\kappa_\theta$), whose corrected increments enter the pose graph. Committed recognitions are registered metrically on the echoes of both visits before they enter the pose graph as loop closure links. The occupancy grid is rendered from the keyframe evidence, placed with the current estimate $\hat{X}$ of the pose graph, so that it follows every loop closure; it does not feed back into the pose estimate.}
\label{fig:overview}
\end{figure*}
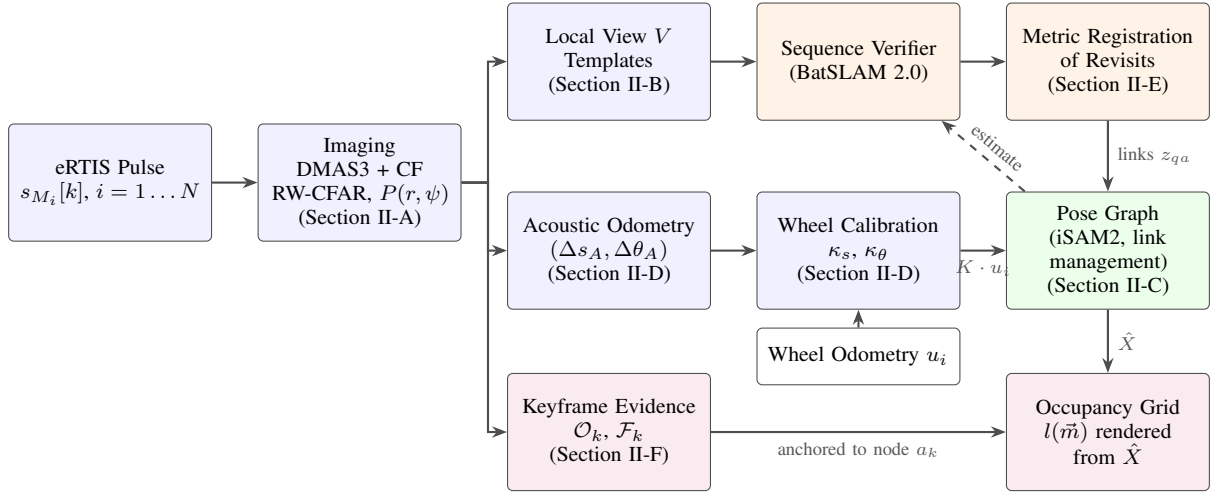

Figure \ref{fig:overview} shows the processing flow of RTISSLAM. For every pulse, the eRTIS records the echoes with its $N = 32$ microphones, which are beamformed with third-order DMAS (Delay-Multiply-And-Sum) with coherence factor (CF) into an energyscape $P(r,\psi)$, followed by a range-wise CFAR filter to reduce the effect of sidelobes (section \ref{sec:imaging}). The wheel odometry adds a node to the pose graph after every \SI{0.1}{\meter} of travel or \SI{6}{\degree} of rotation. The energyscape that was observed for that node is used in four ways. Firstly, it becomes a local view for the place recognition of BatSLAM 2.0, getting stored in the local view database. Secondly, it is correlated with the energyscape of the previous node, which yields the acoustic odometry, used to calibrate the scale factors of the wheel odometry. Thirdly, when the verifier commits to a recognized revisit, the echoes around both visits are registered, which turns the weak loop closure link in the pose graph from BatSLAM2 into a metric loop closure factor (including a true metric pose displacement). Fourthly, at every keyframe, the echoes are converted into occupied and free evidence in the robot frame, stored with the node, and the occupancy grid is rendered from the current estimate of all node poses. The information flows in one direction only: the grid is built from the pose graph, but never feeds back into it as is common in LIDAR based SLAM (section \ref{sec:limitations}).

\subsection{Energyscapes}
\label{sec:imaging}
We consider an eRTIS-type sensor \cite{kerstensERTISFullyEmbedded2019} with one emitter and $N = 32$ microphones in the $YZ$-plane, at nominal positions $\begin{bmatrix} 0 & \bar{p}_y(i) & \bar{p}_z(i) \end{bmatrix}^T$. The emitter sends a broadband FM sweep, and each microphone signal $s_{M_i}[k]$ is processed by a matched filter:
$$
s_{MF_i}[k] = \mathcal{F}^{-1}\{ S_{M_i}(j\omega) \cdot S_b^*(j\omega) \}
$$
with $S_b(j\omega)$ the spectrum of the emitted signal \cite{steckelBroadband3DSonar2013}. For every direction of interest $\psi$, with azimuth $\theta$ and elevation $\varphi$, the signals are delayed by $\tau_i(\psi)$:
$$
\tau_i(\psi)= f_s (\psi_y(i) + \psi_z(i)) / c
$$
with 
$$\psi_y(i) = \sin(\theta)\cos(\varphi)\,\bar{p}_y(i)$$
and
$$\psi_z(i) = \sin(\varphi)\,\bar{p}_z(i)$$
where
$f_s$ is the sample rate and $c$ the speed of sound, which gives
$$x_i[k] = s_{MF_i}[k + \tau_i(\psi)]$$

Delay-And-Sum (DAS), a standard beamforming method, adds these signals. Third-order Delay-Multiply-And-Sum with Coherence Factor (DMAS3 + CF) \cite{Jansen2026DMAS} instead sums the products of all triplets of signed cube roots, and weights the result with the coherence of the array:
\begin{equation}
s^{BF}_{\psi}[k] = \frac{\big(\sum_{i} x_i[k]\big)^2}{N \sum_{i} x_i[k]^2} \sum_{i < j < l} \hat{x}_i[k]\, \hat{x}_j[k]\, \hat{x}_l[k],
\label{eq:dmas}
\end{equation}
with 
$$\hat{x}_i[k] = \operatorname{sgn}(x_i[k])\, |x_i[k]|^{1/3}$$

The triple sum is computed from power sums in $O(N)$ operations, so that the beamformer runs in real time on the GPU of the sensor \cite{Jansen2026DMAS}. The envelope of $s^{BF}_{\psi}[k]$, averaged in power over range cells of \SI{2}{\centi\meter}, forms the energyscape $P(r,\psi)$. As RTISSLAM builds a 2D map, we restrict ourselves to the horizontal plane, with $M = 181$ directions between \SI{-90}{\degree} and \SI{90}{\degree} and 240 range cells from \SI{0.2}{\meter} to \SI{5}{\meter}.

The sidelobes of a strong reflector spread over all directions at its range, and raise the mean energy at that range. We therefore apply a Range-Wise CFAR (RW-CFAR), which is a method of the cell-averaging CFAR family \cite{richardsFundamentalsRadarSignal2014} whose reference cells are all directions at the same range:
\begin{equation}
P^{+}(r,\psi) = \max\left( P(r,\psi) - \alpha \cdot \big(g * \tilde{P}\big)(r),\ 0 \right),
\label{eq:rwcfar}
\end{equation}
with
$$\tilde{P}(r) = \frac{1}{M}\sum_{m=1}^{M} P(r,\psi_m)$$
where
$\alpha = 2$ and $g$ is a Gaussian kernel of three range cells. As $\tilde{P}$ is constant over the directions, this smoothing is equivalent to smoothing the threshold map with a symmetric 2D Gaussian. In what follows, $P_i(r,\psi)$ is the energyscape $P^{+}$ at pose-graph node $i$, and $L_i(r,\psi)$ its range-compensated level in decibels:
$$L_i(r,\psi) = 10\log_{10} P_i(r,\psi) + 40\log_{10} r$$

\subsection{Local Views and Place Recognition}
\label{sec:views}
In BatSLAM 2.0 \cite{steckelBatSLAM20Sequenceverified2026}, places are recognized by comparing local views, i.e., small images that summarize what the sensor perceives at a node, with the views stored as templates along the trajectory. There, the local view was a binaural cochleogram, which is a time-frequency representation from each ear. In this paper, we build this local view template from the energyscape. As a place should be recognized from the layout of its strongest reflectors rather than from the exact shape and strength of individual echoes, the view is made coarse and compressed in four steps.

Firstly, we compensate the range dependence of the echo strength and remove the noise floor. The energy of an echo decays with $r^4$ (spherical spreading on the way out and back), and multiplying by $r^2$ compensates one of these two factors, so that distant reflectors still contribute without dominating the view. As a side note, multiplying with $r^4$ is too aggressive and exagerates noise contributions. The noise floor is removed by subtracting the median of the view:
\begin{equation}
\check{P}_i(r,\psi) = \max\left( r^2 \cdot P_i(r,\psi) - P_{0,i},\ 0 \right),
\end{equation}
where $P_{0,i}$ is the median of $r^2 \cdot P_i(r,\psi)$ over all cells of the energyscape of node $i$.

Secondly, we pool the energyscape into a coarse grid of $A = 20$ azimuth sectors of \SI{9}{\degree} and range bins of \SI{6}{\centi\meter}, up to \SI{4.5}{\meter}:
\begin{equation}
\bar{P}_i[a,b] = \frac{1}{|\mathcal{C}_{ab}|} \sum_{(r,\psi) \in \mathcal{C}_{ab}} \check{P}_i(r,\psi),
\end{equation}
where $\mathcal{C}_{ab}$ is the set of energyscape cells that fall into sector $a$ and range bin $b$. The coarse sectors make the view insensitive to small changes of heading, which change the bearing of every reflector, and a similar point can be made for the range axis.

Thirdly, we smooth the pooled image and compress its dynamic range relative to its strongest bin:
\begin{equation}
V'_i[a,b] = \left( \frac{\big(G * \bar{P}_i\big)[a,b]}{\max_{a,b} \big(G * \bar{P}_i\big)[a,b]} \right)^{\gamma/2},
\label{eq:view}
\end{equation}
where $G$ is a Gaussian kernel with a standard deviation of 0.5 sectors and 0.7 range bins, and $\gamma = 0.2$. As $\bar{P}_i$ is an energy, the exponent $\gamma/2$ raises the echo amplitude to the power $\gamma$. This strong compression keeps weak reflectors visible next to strong specular ones (i.e. walls). Finally, the local view is the zero-mean, unit-norm version of this image:
$$V_i = (V'_i - \mu_i) / \| V'_i - \mu_i \|$$
with $\mu_i$ the mean of $V'_i$. The choice for 20 sectors and $\gamma = 0.2$ were tuned on a separate development drive (section \ref{sec:resultsViews}).

The view keeps range as its second axis, so that the similarity of BatSLAM 2.0 applies without change. The similarity between the view of a query node $q$ and a template $t$ is the highest correlation coefficient over small shifts in range:
\begin{equation}
s(q,t) = \max_{|\delta| \le 3} \rho\big( V_q^{(\delta)}, V_t \big),
\label{eq:similarity}
\end{equation}
where $V_q^{(\delta)}$ is the view of $q$ shifted by $\delta$ range bins and $\rho$ is the correlation coefficient. A revisit seldom happens at exactly the same position: if the robot is slightly further along the corridor than when the template was laid, all reflectors appear closer by the same amount. The winning shift $\delta^*$ therefore also estimates this forward offset, $d^* = \delta^* \cdot \SI{6}{\centi\meter}$, which is used in the loop closure link (section \ref{sec:posegraph}).

Templates are laid every \SI{0.3}{\meter} of travel, and every node is compared with all templates laid more than \SI{5}{\meter} of travel earlier. The best candidates are handed to the sequence verifier of BatSLAM 2.0, which only accepts a recognition when a consistent sequence of candidates supports it. To this end, every candidate contributes evidence according to its similarity:
\begin{equation}
\ell(s) = \operatorname{clip}\big(a \cdot (s - s_0),\ \pm 4\big),
\label{eq:evidence}
\end{equation}
where $s_0$ is the similarity at which a candidate is as likely to be correct as not, and the slope $a$ sets how quickly the evidence grows with the similarity. A hypothesis is committed once its accumulated evidence exceeds the thresholds of BatSLAM 2.0, with stricter thresholds when the commit would correct the map strongly. As these thresholds are expressed in units of $\ell(s)$, the evidence curve has to be fitted for every descriptor. We describe the approach to perform this fit in section \ref{sec:resultsViews}. The plausibility gate and the link management of BatSLAM 2.0 are used without change.

\subsection{The Pose Graph}
\label{sec:posegraph}
Every node $i$ holds the pose $x_i = (x, y, \theta)$ of the robot, connected to the previous node by the wheel odometry increment $u_i$. A committed recognition adds a link between a query node $q$ and the anchor $a$ of the matched template, with measurement $z_{qa}$. With respect to BatSLAM 2.0, the odometry increments are corrected with a calibration matrix $K$, and every link carries its own covariance $\Sigma_{qa}$:
\begin{equation}
\begin{aligned}
\hat{X} = \arg\min_{X} \; & \sum_{i} \big\| \Delta(x_{i-1}, x_i) - K \cdot u_i \big\|^2_{\Sigma_{o,i}} \\
+ & \sum_{(q,a) \in \mathcal{L}} \rho_H\Big( \big\| \Delta(x_q, x_a) - z_{qa} \big\|_{\Sigma_{qa}} \Big),
\end{aligned}
\label{eq:map}
\end{equation}
where $\Delta(x_a, x_b)$ is the pose of node $b$ in the frame of node $a$, $\mathcal{L}$ is the set of links in the graph, $\rho_H$ is the Huber loss and $K = \operatorname{diag}(\kappa_s, \kappa_s, \kappa_\theta)$ (section \ref{sec:calibration}). An unregistered link has $z_{qa} = (d^*, 0, 0)$ and the weak covariance settings of BatSLAM 2.0: (\SI{0.30}{\meter}, \SI{0.20}{\radian}). Equation \ref{eq:map} is solved incrementally with iSAM2 \cite{kaessISAM2IncrementalSmoothing2012} in GTSAM \cite{dellaertFactorGraphsGTSAM2012}, and the links of every committed hypothesis form a group that the link management of BatSLAM 2.0 can withdraw and re-admit.

\subsection{Acoustic Odometry and Wheel Calibration}
\label{sec:ao}
\label{sec:calibration}
A static reflector at range $r$ and bearing $\theta$ moves through the image of a sensor that moves with linear velocity $V$ and angular velocity $\omega$ according to the acoustic flow \cite{steckelBroadband3DSonar2013, steckelAcousticFlowBasedControl2017}: $\dot{r} = -V\cos(\theta)$ and $\dot{\theta} = V\sin(\theta)/r - \omega$. In \cite{steckelBroadband3DSonar2013}, consecutive energyscapes were approximated as translated in $(r,\theta)$, which underestimates the forward motion, as reflectors off the axis approach more slowly than the sensor moves. We therefore correlate after an exact warp: a point $\vec{p}$ in the current sensor frame lies, in the previous one, at
\begin{equation}
\vec{p}\,' = R(\Delta\theta) \cdot (\vec{p} + \vec{m}) + \vec{t}(\Delta s, \Delta\theta) - \vec{m},
\label{eq:warp}
\end{equation}
with $R(\cdot)$ the rotation matrix, $\vec{m}$ the mount of the sensor (\SI{0.1}{\meter} in front of the rotation center) and $\vec{t} = \Delta s \begin{bmatrix} \cos(\Delta\theta/2) & \sin(\Delta\theta/2) \end{bmatrix}^T$ the chord of a circular arc. With $\mathcal{W}(\Delta s, \Delta\theta)$ the operator that resamples an image at these positions, the acoustic odometry between nodes $i-1$ and $i$ is
\begin{equation}
\big(\Delta s_{A,i}, \Delta\theta_{A,i}\big) = \operatorname*{arg\,max}_{(\Delta s, \Delta\theta) \in \mathcal{S}_i} \rho\Big( \breve{P}_i, \mathcal{W}(\Delta s, \Delta\theta)\, \breve{P}_{i-1} \Big),
\label{eq:ao}
\end{equation}
where $\breve{P}_i = \min\big(1, \max(0, (L_i - \max L_i + 25)/25)\big)$ is the level in a \SI{25}{\deci\bel} window below the peak, between \SI{1.5}{\meter} and \SI{4.5}{\meter} and within \SI{\pm 80}{\degree}, and $\mathcal{S}_i$ is a window of $\pm 4$ range cells and \SI{\pm 5}{\degree} around the wheel increment, searched coarse to fine with a parabolic peak fit. Estimates whose correlation peak is not clearly above the median of the window are rejected.

Using equation \ref{eq:ao} as a second odometry factor would stiffen the odometry chain with the small, environment-dependent biases of the acoustic odometry. The errors of the wheels, on the other hand, are dominated by a scale error on the distance and a heading bias, which are hard to observe from the wheels but easy to observe from the environment. We therefore let the acoustic odometry calibrate the wheels, with the least-squares scale factors
\begin{equation}
\kappa_s = \frac{\sum_i \Delta s_{A,i}\, \Delta s_{W,i}}{\sum_i \Delta s_{W,i}^2}, \qquad \kappa_\theta = \frac{\sum_i \Delta \theta_{A,i}\, \Delta \theta_{W,i}}{\sum_i \Delta \theta_{W,i}^2},
\label{eq:calib}
\end{equation}
with $\Delta s_{W,i}$ and $\Delta\theta_{W,i}$ the wheel increments, over the pairs with a sharp correlation peak (a factor 1.6 above the median). $K$ is applied once 40 pairs and \SI{0.5}{\radian\squared} of rotation have been accumulated, and the search window stays centered on the raw wheel increment, as a broad correlation peak would otherwise lean towards a calibrated prediction and confirm it.

\subsection{Metric Registration of Recognized Revisits}
\label{sec:registration}
A link of BatSLAM 2.0 places the robot at the anchor pose, shifted along its heading, with a large covariance, as a binaural view says little about a lateral offset. An imaging sonar observes where its reflectors are, so that a revisit can be registered. For a committed pair $(q, a)$, the echo detections (section \ref{sec:grid}) of the nodes within \SI{1.2}{\meter} of travel before $q$ form a point set $\mathcal{Q}$ in the frame of $q$, and those within \SI{\pm 1.2}{\meter} around $a$ a set $\mathcal{A}$ in the frame of $a$, both composed with odometry. Following correlative scan matching \cite{olsonProbabilisticSelfLocalization2000}, a relative pose $z$ is scored by
\begin{equation}
S(z) = \frac{1}{|\mathcal{Q}|} \sum_{\vec{p} \in \mathcal{Q}} \min\Big(1, \sum_{\vec{p}\,' \in \mathcal{A}} e^{-\| T(z)^{-1}\vec{p} - \vec{p}\,' \|^2 / 2\sigma^2} \Big),
\label{eq:registration}
\end{equation}
with $T(z)$ the rigid transformation of $z$ and $\sigma = \SI{5}{\centi\meter}$, maximized over \SI{\pm 0.5}{\meter} and \SI{\pm 15}{\degree} around $(d^*, 0, 0)$. If the maximum is high ($S \ge 0.25$) and distinct (at least twice the median score of the search), it replaces $z_{qa}$, with a tight covariance (\SI{6}{\centi\meter}, \SI{0.03}{\radian}) - otherwise the link keeps its uncertain BatSLAM2.0 form.

\subsection{Occupancy Grid Anchored to the Pose Graph}
\label{sec:grid}
The occupancy grid is not built from every pulse, but from keyframes, which are taken every \SI{0.25}{\meter} of travel or \SI{15}{\degree} of rotation. Every keyframe is anchored to the pose-graph node at which it was taken, and stores its evidence in the frame of the robot, i.e., independent of where the pose graph currently places that node. In what follows, we first describe which evidence a keyframe stores, and then how the grid is rendered from it.

A cell of the energyscape is taken as a detection when its range-compensated level exceeds a threshold that adapts to the content of the view:
\begin{equation}
T_k = \max\left( \tilde{L}_k + \SI{22}{\deci\bel},\ \hat{L}_k - \SI{30}{\deci\bel} \right),
\label{eq:detection}
\end{equation}
where $\tilde{L}_k$ is the median level and $\hat{L}_k$ the peak level of the energyscape of keyframe $k$. The first term keeps the threshold well above the noise floor of a sparse view, while the second one prevents the sidelobes of a very strong echo from being detected in a crowded one. Every detection becomes a point of occupied evidence, at range $r$ and direction $\psi$ as seen from the sensor. Each point is weighted by how far its level exceeds the threshold, from 0.4 at the threshold to 1 at \SI{10}{\deci\bel} above it, so that strong echoes count more than marginal ones.

Free evidence is laid along every direction, from the sensor up to \SI{12}{\centi\meter} before the first detection in that direction, with at most \SI{3.5}{\meter}. In directions without any detection, however, free evidence is only laid up to \SI{0.6}{\meter}. This rule is specific to sonar: a specular wall at grazing incidence returns no echo at all, so that the absence of an echo does not mean that the space is free \cite{moravecHighResolutionMaps1985}. Without it, every wall that is passed at a shallow angle would be erased from the map.

The grid is rendered from the evidence of all keyframes, each placed with the current pose estimate of its anchor node. For every cell $\vec{g}$ of the grid, we count the occupied and free points that fall into it, and combine them into log-odds \cite{thrunProbabilisticRobotics2005}:
\begin{equation}
l(\vec{g}) = \operatorname{clip}\Big( l_o \cdot N_o(\vec{g}) - l_f \cdot N_f(\vec{g}),\ -3,\ 6 \Big),
\label{eq:grid}
\end{equation}
where $N_o(\vec{g})$ is the weighted number of occupied points and $N_f(\vec{g})$ the number of free points in cell $\vec{g}$, over all keyframes, and $l_o = 0.9$ and $l_f = 0.25$ are the log-odds increments of a single occupied and free point. The clipping keeps every cell able to change its state when new evidence arrives.

As the evidence is stored per keyframe in the frame of the robot, the grid is not updated incrementally, but rendered as a whole whenever the pose graph changes, which takes \SI{0.16}{\second} for 739 keyframes with \SI{5}{\centi\meter} cells. A loop closure that corrects the pose graph therefore moves the evidence of every keyframe along with its anchor node, and a group of links that is withdrawn by the link management takes its effect on the map with it. This is the main difference with grid-based SLAM systems that write evidence into a single global grid \cite{grisettiImprovedTechniquesGrid2007} or into fixed submaps \cite{hessRealtimeLoopClosure2016}, in which a correction of the trajectory cannot be propagated into evidence that has already been written.

\section{Experimental Setup}
\label{sec:setup}
We evaluated RTISSLAM in a closed-loop simulator, in which a differential-drive robot with an eRTIS drives through a floor plan while the sonar is simulated pulse by pulse.

\begin{figure*}[t]
\centering
\includegraphics[width=\linewidth]{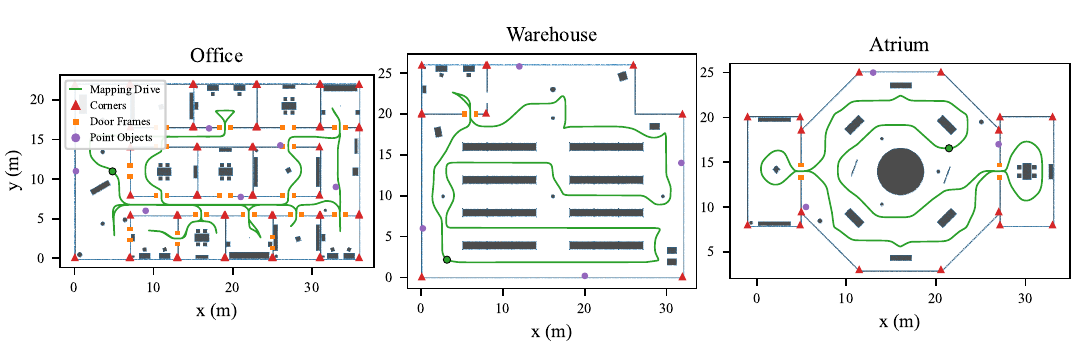}
\caption{The Three Simulated Worlds: the office, the warehouse and the atrium. Grey: walls and furniture footprints; blue: reflector points at sensor height; red triangles: concave corners; orange squares: door frames; purple: point objects; green: the mapping tour, starting at the circle.}
\label{fig:worlds}
\end{figure*}

\textbf{Worlds:} We built three floor plans with specular walls and few point reflectors (figure \ref{fig:worlds}), as these are the conditions in which sonar mapping is hardest. The office ($36 \times \SI{22}{\meter}$) has a lobby with pillars, two corridors and 13 furnished rooms, with loops at very different scales. The warehouse ($32 \times \SI{26}{\meter}$, L-shaped) has eight double-sided rack rows, a cross aisle and a walled office. Its aisles are nearly identical, which makes it the most aliased world. The atrium is an octagonal hall (radius \SI{12}{\meter}) around a round glass core, with planters, glass screens and two side rooms. This world is the most open world of the project. Walls are occluding boxes whose faces carry reflector points with a specular backscatter lobe, door openings carry door frames, every concave wall junction carries a corner reflector, and furniture (desks, cabinets, racks, tables and chairs with thin legs) is modeled with occluding boxes and cylinders.

\begin{figure*}[t]
\centering
\includegraphics[width=\linewidth]{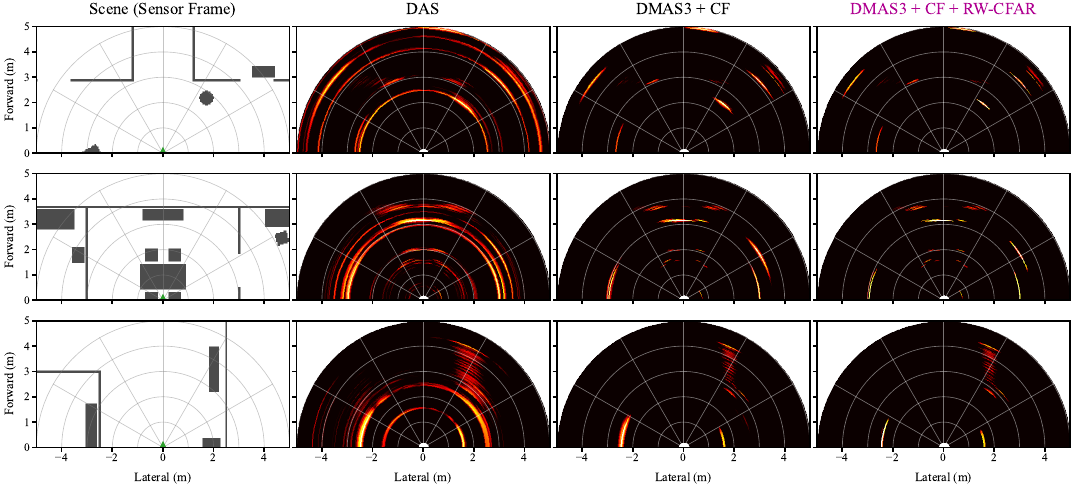}
\caption{Energyscapes at Three Poses in the Office, in the sensor frame (forward up), range-compensated ($+40\log_{10} r$) and normalized per image, with the colormap \raisebox{-0.15ex}{\includegraphics[height=1.4ex,width=1.6cm]{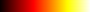}} from \SI{-30}{\deci\bel} to \SI{0}{\deci\bel}. Left: the scene; then Delay-And-Sum (DAS), third-order Delay-Multiply-And-Sum with Coherence Factor (DMAS3 + CF), and DMAS3 + CF with the Range-Wise CFAR used by RTISSLAM. Grid lines every \SI{30}{\degree} and \SI{1}{\meter}.}
\label{fig:energyscapes}
\end{figure*}

\textbf{Sensor:} The eRTIS is simulated with the geometric, coherent echo synthesis paradigm of SonoTraceLab and SonoTraceUE \cite{jansenSonoTraceLabRaytracingBasedAcoustic2024, jansenHardwareAcceleratedGeometricalSimulation2026}, which has been used to develop and validate sonar-based control and imaging before \cite{jansenRealTimeSonarFusion2022, steckelSelfSupervisedDeconvolution2026}: perturbed microphone positions with gain and phase errors, a band-pass emitter with a misaligned, frequency-dependent directivity, spherical spreading, atmospheric absorption, occlusion and sensor noise. Figure \ref{fig:energyscapes} shows the energyscapes at three poses: walls are only seen near normal incidence, and door frames, corners and the edges of furniture are the dominant landmarks. DMAS3 + CF removes most of the sidelobes and range rings of DAS, and the RW-CFAR the remaining ones.

\textbf{Robot and drives:} The robot has a radius of \SI{0.25}{\meter}, with the sensor \SI{0.1}{\meter} in front of the axle and \SI{0.4}{\meter} above the floor. Its wheel odometry has a scale error of 2~\% on the traveled distance, a heading bias of 3~\% of the rotation, as in BatSLAM 2.0 \cite{steckelBatSLAM20Sequenceverified2026}, and additive white Gaussian noise. The mapping drives follow a tour that visits most rooms and closes loops at several scales, driven twice (2.1 laps) with pure pursuit at up to \SI{0.4}{\meter\per\second} and a pulse rate of \SI{10}{\hertz}: \SI{404}{\meter} in the office (3992 nodes), \SI{330}{\meter} in the warehouse (2988 nodes) and \SI{271}{\meter} in the atrium (2450 nodes). Every drive was repeated with five seeds, which change the sensor and odometry noise and the reflector sampling. The descriptor and its evidence curve were tuned on a separate single-lap office drive (the development drive), and ground truth is used for evaluation only.

\textbf{Baseline:} As a lidar-style alternative, we ran slam\_toolbox \cite{macenskiSLAMToolboxSLAM2021}, i.e., Karto's correlative scan matching with a pose graph and geometric loop closure, on the same data: a pseudo-scan with, per direction, the range of the first detection of equation \ref{eq:detection}, i.e., the same echoes as our grid, and the same wheel odometry. We report its stock parameters, and the best of 38 parameter sets tuned on the development drive.

\section{Results}
\label{sec:results}

\subsection{Place Recognition and the Evidence Curve}
\label{sec:resultsViews}
The local view was tuned on the development drive using templates every \SI{0.3}{\meter}, and for every node the best candidate among the templates laid more than \SI{5}{\meter} of travel earlier. A candidate is correct if its anchor lies within \SI{0.6}{\meter} and \SI{26}{\degree} of the query. With the settings of BatSLAM 2.0 (60 sectors, $\gamma = 0.33$), 34~\% of the revisit nodes are recognized at a precision of 95~\%; coarser sectors and a stronger compression, which make the view less sensitive to the exact shape of the specular echoes, raise this to 54~\% (20 sectors, $\gamma = 0.2$, with an AUC of 0.97). The evidence curve was then fitted in the way of BatSLAM 2.0 on the five best candidates of every node: a monotone (isotonic) regression of the fraction of correct candidates against $s$ gives $s_0 = 0.89$, where half of them are correct, and the slope $a = 49$ follows from where the logit of that fraction reaches $+2$, i.e., from the upper branch of the curve, where true revisits produce their candidates. The candidate floor is set where $\ell(s)$ reaches $-4$ (0.807). This fit matters: with a slope taken from the whole curve (about 30), true revisits never accumulate the evidence that the risk-scaled commit rule requires for large corrections, and the first loop closure is never made.

\subsection{SLAM Accuracy and Ablation}
\label{sec:resultsSlam}
\label{sec:resultsAblation}
\begin{figure*}[t]
\centering
\includegraphics[width=\linewidth]{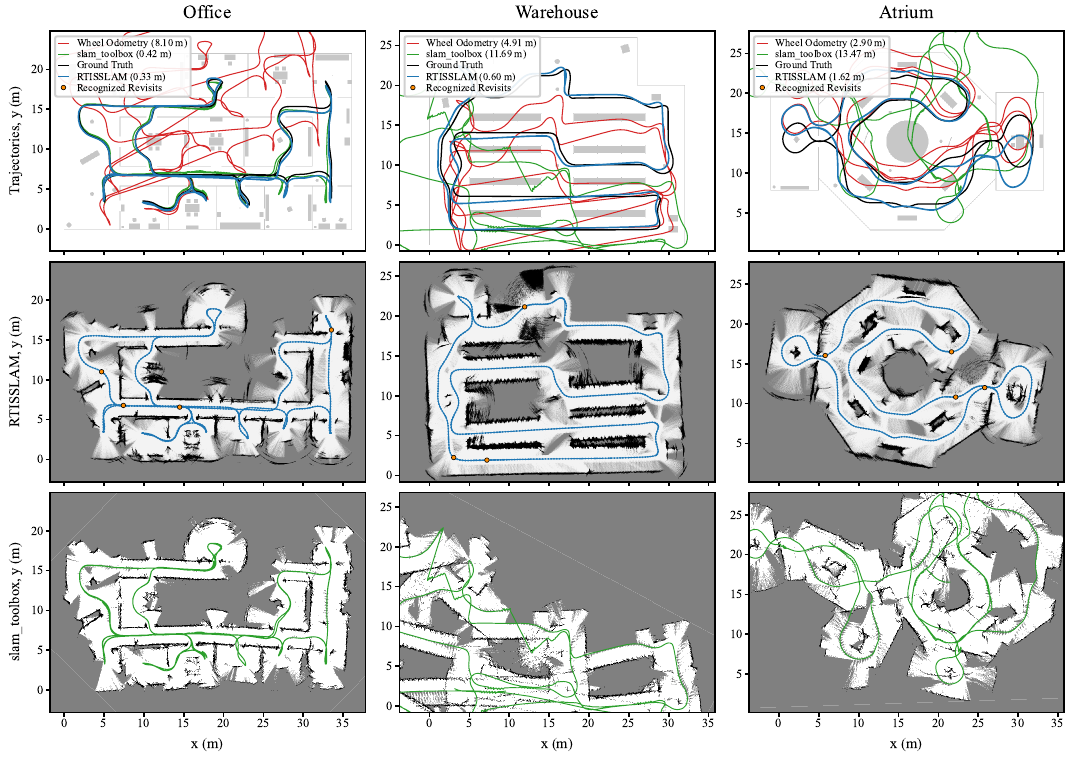}
\caption{RTISSLAM and slam\_toolbox on the Three Worlds (First Seed). Top: ground truth (black), wheel odometry (red), slam\_toolbox (green) and RTISSLAM (blue), with their ATE, on the floor plan (reference only). Middle: the RTISSLAM occupancy grid, rendered from the final pose graph, with the estimated trajectory, its template nodes (blue dots) and the nodes at which a revisit was recognized (orange). Bottom: the slam\_toolbox grid and trajectory on the same sonar echoes and odometry. Grids: black is occupied, white is free and grey is unknown.}
\label{fig:slamWorlds}
\end{figure*}

Table \ref{tab:results} summarizes the Absolute Trajectory Error (ATE) over the five seeds, and figure \ref{fig:slamWorlds} shows the result of the first seed. RTISSLAM reduces the error of the wheel odometry from \SI{8.2}{\meter} to \SI{0.31}{\meter} in the office and from \SI{5.1}{\meter} to \SI{0.57}{\meter} in the warehouse. The online estimate, i.e., the pose of the current node at the moment it is added, is \SI{0.79}{\meter} and \SI{1.13}{\meter}. Out of on average 2090 and 1817 loop closure links per run, none and 14 over all five seeds (0.15~\%) connected places more than \SI{1}{\meter} apart, all in the nearly identical aisles of the warehouse; the robust kernel and the link management keep them from distorting the map.

\begin{figure}[t]
\centering
\includegraphics[width=\linewidth]{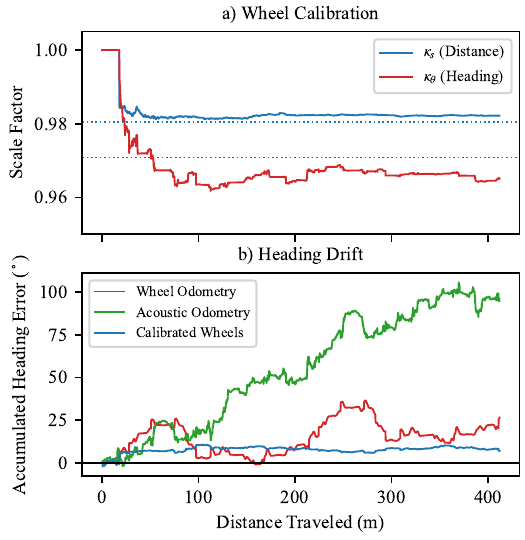}
\caption{Acoustic Odometry on the Office Drive (First Seed). a) Convergence of the wheel calibration (dotted: the true correction factors). b) Accumulated heading error of the raw wheel odometry, the acoustic odometry (wheel increments where an estimate was rejected) and the calibrated wheel odometry.}
\label{fig:ao}
\end{figure}

The ablation shows how the elements contribute. The links of BatSLAM 2.0 alone already reduce the error by an order of magnitude, but without registration and calibration they include 8 false links in the office and 18 in the warehouse over the five seeds, and none in the atrium. Their metric registration helps where the geometry is distinctive (the office and the atrium, where 85~\% and 66~\% of the committed links are registered), and less in the warehouse (38~\%). Without loop closures, the calibration of the wheels by the acoustic odometry reduces the dead-reckoning error by a factor of 3.5 in both indoor worlds, and on top of the links it gives the best result, a further 25~\% and 37~\%. Figure \ref{fig:ao} shows why: the calibration converges within 1~\% of the true correction factors after \SI{23}{\meter} of travel in the office, after which the calibrated wheels combine the low noise of the wheels with the absence of bias of the acoustic odometry, whereas the acoustic odometry by itself performs a random walk in heading. In the atrium, however, the calibration fails: most reflectors are far away and peripheral, only 70~\% of the node pairs yield an acoustic odometry estimate, and the estimated heading scale is biased (0.936 against the true 0.971), so that the calibrated dead reckoning is worse than the raw wheels, and RTISSLAM (\SI{1.78}{\meter}) is worse than the registered links without calibration (\SI{0.53}{\meter}). We come back to this in section \ref{sec:limitations}.

The baseline confirms the premise of RTISSLAM. With its stock parameters, slam\_toolbox is worse than dead reckoning in every world, as Karto's scan matching turns the sparse, specular sonar scans into heading errors. Tuned on the development drive, it reaches \SI{0.42}{\meter} on one office drive, but diverges on another (\SI{23.7}{\meter}), and it does not transfer: in the warehouse and the atrium, it is worse than dead reckoning on all ten drives, as scan matching slides along repetitive aisles and open space. Its loop closures rarely fire, so that its error is that of the scan-to-scan matching. The bottom rows of figure \ref{fig:slamWorlds} show the consequence for the maps: the RTISSLAM grids resolve walls, doors, racks and furniture, whereas the slam\_toolbox grids are smeared and doubled.

\begin{table}[t]
\centering
\caption{Absolute Trajectory Error After the Two-Lap Drives (m, Mean$\,\pm\,$Standard Deviation Over Five Seeds).}
\label{tab:results}
\setlength{\tabcolsep}{3pt}
\footnotesize
\begin{tabular}{lccc}
\toprule
Variant & Office & Warehouse & Atrium \\
\midrule
Wheel odometry & 8.20$\,\pm\,$0.33 & 5.09$\,\pm\,$0.35 & 2.66$\,\pm\,$0.34 \\
\quad + AO calibration & 2.32$\,\pm\,$0.47 & 1.48$\,\pm\,$0.25 & 2.99$\,\pm\,$0.61 \\
\midrule
BatSLAM 2.0 links & 0.46$\,\pm\,$0.06 & 0.82$\,\pm\,$0.10 & 1.03$\,\pm\,$0.07 \\
\quad + registration & 0.41$\,\pm\,$0.05 & 0.91$\,\pm\,$0.12 & \textbf{0.53$\,\pm\,$0.10} \\
\quad + AO calib.\ (RTISSLAM) & \textbf{0.31$\,\pm\,$0.13} & \textbf{0.57$\,\pm\,$0.13} & 1.78$\,\pm\,$0.32 \\
\midrule
slam\_toolbox, stock & 23.1$\,\pm\,$8.2 & 35.5$\,\pm\,$9.4 & 17.3$\,\pm\,$2.4 \\
slam\_toolbox, tuned & 6.1$\,\pm\,$9.9 & 15.6$\,\pm\,$5.3 & 14.9$\,\pm\,$2.4 \\
\bottomrule
\end{tabular}
\end{table}

\subsection{The Occupancy Grid}
\label{sec:resultsGrids}
\label{sec:resultsLoop}
\begin{figure}[t]
\centering
\includegraphics[width=\linewidth]{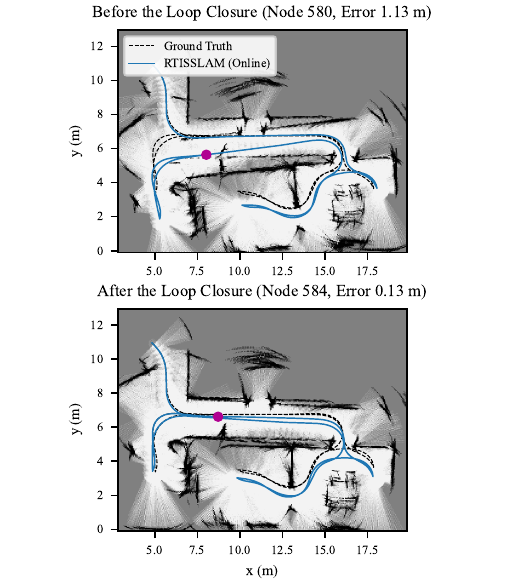}
\caption{The Online Occupancy Grid of the Office Just Before and After a Loop Closure. Dashed black: ground truth (reference only); blue: online RTISSLAM estimate; magenta: current node. The keyframes are placed with the current estimate of their anchor nodes, so that the loop closure moves their evidence with them.}
\label{fig:loopClosure}
\end{figure}

\begin{figure*}[t]
\centering
\includegraphics[width=\linewidth]{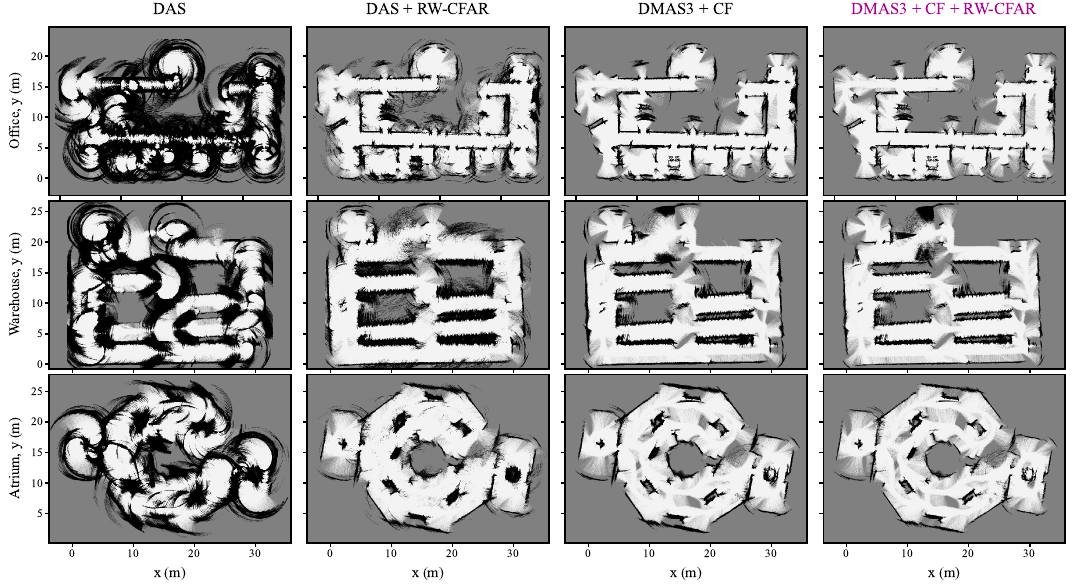}
\caption{Occupancy Grids per Imaging Method in the Three Worlds, rendered from the final RTISSLAM poses (first seed) with the same inverse sensor model: Delay-And-Sum (DAS) and DMAS3 + CF, each without and with the Range-Wise CFAR. The configuration of RTISSLAM is highlighted.}
\label{fig:grids}
\end{figure*}

Figure \ref{fig:loopClosure} shows the online grid of the office just before and after a loop closure, at node 584, which reduces the error of the current node from \SI{1.13}{\meter} to \SI{0.13}{\meter} and moves earlier nodes by up to \SI{1.6}{\meter}. Before the closure, the walls of the revisited part appear twice; after it, the keyframes of both visits are placed with their corrected anchors, and the double walls merge without a trace of the erroneous placement. Figure \ref{fig:grids} compares the grids of DAS and DMAS3 + CF, with and without the RW-CFAR, in the three worlds, rendered from the same poses with the same inverse sensor model: without the RW-CFAR, the sidelobes and range rings of DAS are taken for obstacles, and the RW-CFAR removes most of them even for DAS; DMAS3 + CF with the RW-CFAR gives the cleanest walls and the fewest phantom detections.

\section{Discussion and Limitations}
\label{sec:limitations}
An argument one could validly bring up is that RTISSLAM does not use its own map to its full extend: the occupancy grid is built from the pose graph, but it never constrains the pose estimate, whereas the accuracy of lidar SLAM stems precisely from matching scans against the occupancy map \cite{grisettiImprovedTechniquesGrid2007, hessRealtimeLoopClosure2016}. This choice was made deliberately based on the observations shown in the baseline results: a single sonar scan is sparse, specular and view dependent, and matching it against the scans or the grid of earlier visits drives slam\_toolbox into errors larger than those of the wheels. The raytracing approach that makes LIDAR scan matching successfull in occupancy grid SLAM does not transfer to sonar sensing.  By keeping the grid out of the estimator, the sequence verification and link management of BatSLAM 2.0 remain the controlling factor of the loop closures, rendering the grid consistent with the pose graph after every optimization. The same graph can render grids from other imaging methods afterwards (figure \ref{fig:grids}). The metric information of the sonar that is present in the RTISSLAM system is utilized, as it enters through the acoustic odometry and the registration of recognized revisits, which are local comparisons in which the artifacts of the physics driving sonar sensing are sufficiently small. The price is that geometric constraints between places that are not recognized as the same place are lost, such as a corridor driven in the opposite direction. Therefore, registering keyframes that the graph places close together, under the same link management, would be a natural extension.

An additional limitation of this paper can be found in the simulation-centered approach. The simulator follows a validated paradigm \cite{jansenSonoTraceLabRaytracingBasedAcoustic2024, jansenHardwareAcceleratedGeometricalSimulation2026}, and the worlds were made deliberately hard, but floors, ceilings and multiple reflections are not modeled in our 2D slice, so that a real-world evaluation with a lidar reference is the obvious next step. Secondly, the calibration of the wheels relies on what is assumed to be an unbiased acoustic odometry method. In the open atrium, the heading scale of the acoustic odometry is actually biased due to the geometry of the scene, and the calibration then makes the map worse instead of better. Improvements can be found in a calibration method that is only applied when enough well-conditioned pairs support it, or that is estimated jointly in the pose graph. Thirdly, the descriptor that recognizes most revisits also produces a few false loop closures in strongly aliased environments. The robust back-end absorbed them in all our runs, but they remain a risk (as is a generally accepted risk in all SLAM systems that perform explicit loop closures). Finally, our wheel odometry has a constant bias, whereas the calibration of a real robot may have to follow slow changes of load, tire pressure or floor. The acoustic-odometry based calibration can naturally be extended to a continuous estimate of the calibration matrix.

\section{Conclusion}
\label{sec:conclusion}
In this letter, we presented RTISSLAM, a novel sonar-only SLAM system for an embedded 3D imaging sonar, which combines the robust pose graph of BatSLAM 2.0 as back-end with an occupancy grid that is anchored to the pose-graph nodes. Energyscapes from DMAS beamforming with a range-wise CFAR serve as place descriptors for sequence-verified loop closures, acoustic odometry calibrates the wheel odometry online, and recognized revisits are registered metrically, without any learned component. Over two laps through an office and a warehouse with specular walls, RTISSLAM reduced the trajectory error of the wheel odometry from \SI{8.2}{\meter} and \SI{5.1}{\meter} to \SI{0.31}{\meter} and \SI{0.57}{\meter} (five seeds), and built grids in which walls, doors and furniture are clearly resolved, whereas a lidar-style grid SLAM system on the same sonar data did worse than the wheels in most runs. In an open atrium, the loop closures and their registration reduced the error to \SI{0.53}{\meter}, but the calibration from acoustic odometry did not hold. In future work, we will validate RTISSLAM on a real robot with a lidar reference, make the calibration robust in open spaces. We have implemented a ROS 2 version of the full algorithm, compatible with the Nav2 stack, which is demonstrated in the accompanying video\ifanonymous\else{} (\url{https://youtu.be/OV5ceQ0kd-c})\fi. The full code is available at \url{https://github.com/Cosys-Lab/RTISSLAM}.

\section*{Acknowledgment}
In accordance with the IEEE policy on the use of artificial intelligence, we disclose that the AI assistant Claude (Anthropic) was used in the preparation of this work: for the implementation of the simulation and SLAM software, for running the experiments, for generating the figures and the supplementary video, and for drafting parts of the text. All results were checked, and the text was reviewed and edited, by the \ifanonymous authors\else author\fi, who takes full responsibility for the content of this article.

\bibliographystyle{IEEEtran}
\bibliography{references,in_air_sonar_2024_2026}

\end{document}